\documentclass[letterpaper, 10 pt, conference]{ieeeconf}  

\usepackage{times}
\usepackage[T1]{fontenc}
\usepackage{graphicx}
\makeatletter
\let\NAT@parse\undefined
\makeatother
\usepackage[numbers]{natbib}
\usepackage{multicol}
\usepackage{booktabs} 
\usepackage[bookmarks=true]{hyperref}
\usepackage{cleveref}

\usepackage{xcolor}
\usepackage{verbatim}
\def\ourhand#1{HIRO Hand {#1}}
\IEEEoverridecommandlockouts

\title{\LARGE \bf
A Wearable Robotic Hand for Hand-over-Hand Imitation Learning
}

\author{Dehao Wei$^{1}$ and Huazhe Xu$^{2}$
\thanks{$^{1}$Dehao Wei is with the Tsinghua-Berkeley Shenzhen Institute, Tsinghua University,
        BeiJing, CN 100000, CHINA
        {\tt\small weidh21@mails.tsinghua.edu.cn}}
\thanks{$^{2}$Huazhe Xu is with the Institute for Interdisciplinary Information Sciences, Tsinghua University, Shanghai AI Lab, and Shanghai Qi Zhi Institute
        {\tt\small huazhe\_xu@mail.tsinghua.edu.cn}}
}
\begin{document}
\maketitle
\pagestyle{empty}  
\thispagestyle{empty} 
\thispagestyle{empty}
\pagestyle{empty}
\begin{abstract}
Dexterous manipulation through imitation learning has gained significant attention in robotics research. 
The collection of high-quality expert data holds paramount importance when using imitation learning.
The existing approaches for acquiring expert data commonly involve utilizing a data glove to capture hand motion information. However, this method suffers from limitations as the collected information cannot be directly mapped to the robotic hand due to discrepancies in their degrees of freedom or structures. Furthermore, it fails to accurately capture force feedback information between the hand and objects during the demonstration process. 
To overcome these challenges, this paper presents a novel solution in the form of a wearable dexterous hand, namely \underline{H}and-over-hand \underline{I}mitation learning wearable \underline{R}\underline{O}botic Hand~(HIRO Hand), which integrates expert data collection and enables the implementation of dexterous operations.
This \ourhand empowers the operator to utilize their own tactile feedback to determine appropriate force, position, and actions, resulting in more accurate imitation of the expert's actions. We develop both non-learning and visual behavior cloning based controllers allowing  \ourhand successfully achieves grasping and in-hand manipulation ability. More details are accessible through the link: \href{https://sites.google.com/view/hiro-hand/%E9%A6%96%E9%A1%B5}{Project website}
\end{abstract}
\section{INTRODUCTION}
\begin{figure}[t!]
    \centering\includegraphics[width=8.7cm]{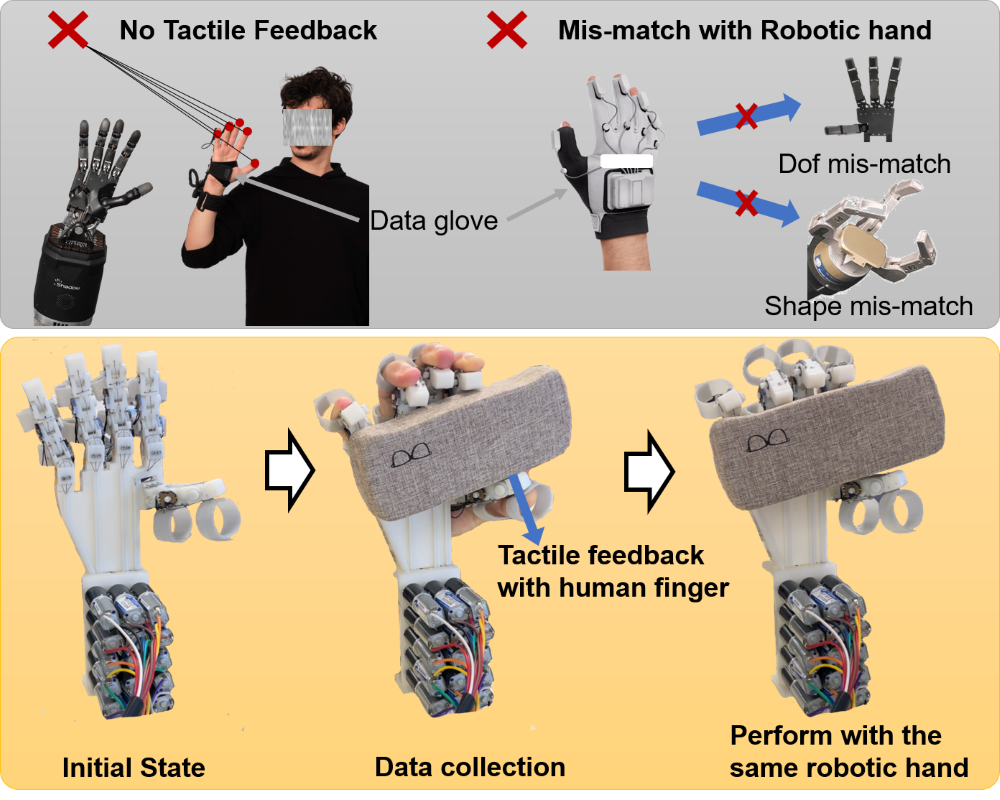}
    \caption{Overview of the HIRO Hand. The \ourhand addresses two main challenges encountered in traditional expert data collection methods: 1) the lack of sensing feedback from objects, and 2) the mismatch between data gloves and the robotic hand. The \ourhand serves a dual purpose by enabling data collection through wearing and implementing the learned manipulation abilities.}
    \label{fig1}
\end{figure}

Dexterous manipulation within a learning paradigm has gained significant attention in recent years~\cite{dexterous_by_rl_1, dexterous_by_rl_2,dexterous_by_rl_3,overview_of_dexterous_manipulation}. 
Recently, dexterous manipulation based on imitation learning has become increasingly popular~\cite{imitation_learning_review, dexterous_by_il_1}. 
One of the keys to achieving dexterous manipulation through imitation learning lies in obtaining high-quality expert data. 
To address this, researchers have explored new paradigms using demonstration data from various sources such as data gloves~\cite{sensing_glove}, virtual reality~\cite{8461249}, and videos~\cite{Sivakumar_2022}. Although these methods offer advantages in terms of data efficiency and action naturalness, they have drawbacks related to contact information feedback and mismatches between the data glove and the robotic hand. 
In this paper, we propose a novel approach by introducing the HIRO Hand, which integrates a dexterous hand with a data collection device. Figure \ref{fig1} illustrates the hand-over-hand data collection process, where human users can utilize self-finger sensing to detect contact information. Moreover, the HIRO Hand exhibits high-performance manipulation capabilities.
The contributions of this paper can be summarized as follows:
\textbf{1)} A novel wearable dexterous hand, integrating expert data collection and dexterous manipulation, is proposed to addresses the tactile feedback limitations when collecting expert data using data gloves.
\textbf{2)} Controllers based on PID and visual imitation learning were developed, respectively, and enabled the \ourhand to demonstrate more than 10 different grasping and manipulation tasks.
\textbf{3)} We develop a fully 3D-printed, 15-degree-of-freedom (DOF) dexterous hand, which is cost-effective (400 dollars) and demonstrates a repeat deviation lower than 0.14 mm. This hand is capable of handling 0.80 of human grasp types~\cite{Cini_2019}.

\section{RELATED WORK}
\subsection{Imitation Learning with Dexterous Hands}
Imitation learning~(IL) is a widely utilized technique for achieving dexterous hand manipulation~\cite{Sivakumar_2022,Mandikal2022DexVIPLD,Song_2020,8461249,li2021igibson,sensing_glove,7002126}. However, an outstanding challenge is the acquisition of expert data. 
Numerous algorithms~\cite{Sivakumar_2022, Mandikal2022DexVIPLD} have been developed to obtain data from videos, either through expert demonstrations or the internet. While internet data is cost-effective and enables fast acquisition of examples, the quality of the data acquired is usually unsatisfactory. 
Collecting data through virtual platforms~\cite{8461249,li2021igibson} or data gloves~\cite{sensing_glove,7002126} may improve data quality. Nonetheless, the lack of sensory feedback mechanisms (e.g., contact force, fingertip tactile) in these systems leads to imperfect demonstrations, and data collection platforms are relatively expensive. 
In this research, the proposed \ourhand leverages the contact physics of human fingers during hand-over-hand teaching to enable more precise operations, resulting in the collection of data that is both cost-effective and of high quality.
\subsection{Anthropomorphic Robotic Hands}
Numerous dexterous anthropomorphic robotic hands have been developed. Tendon-driven mechanisms are the most similar to the human hand structure, where tendons connect to actuators located on the forearm and transmit the driving force to the robotic hand joints~\cite{18,19,20}. In this study, we concentrate on tendon-driven multi-DOFs hands with high dexterity\citep{10}.
 The FLLEX hand~\cite{18} exhibits high impact-absorbing performance and force output. Nonetheless, the use of tendons to convey the driving force may cause a reduction in joint control precision due to their deformation. Moreover, the FLLEX hand integrates 15 servo motors, which may incur a relatively high cost. Another example of tendon-driven hands is the biomimetic robotic hands~\cite{19} reinterpret important biomechanical advantages of the human hand from a roboticist's perspective and closely mimic the human counterpart. However, this robotic hand uses ten Dynamixel servos, which may not be able to adapt to large forces and cannot provide precise control of the finger. 
The Shadow Dexterous Hand, developed by Shadow Robot Company, is a high-DOFs tendon-driven robotic hand that demonstrates good manipulation performance~\cite{21}. However, the product is expensive, and the maintenance cost is high due to the complexity of the hand's components. In general, existing tendon-driven robotic hands demonstrate a high degree of dexterity; however, their accessibility is limited by their relatively high cost. Besides, The lack of ability to collect expert data necessitates reliance on external devices or algorithm that may result in lower quality expert data. The proposed robotic hand presented offers a low-cost alternative, with high dexterity.
\section{ROBOT DESIGN}
We present the mechanical design and the forward/inverse kinematic model of the \ourhand.

\begin{figure}[ht]
    \centering\includegraphics[width=8.5cm]{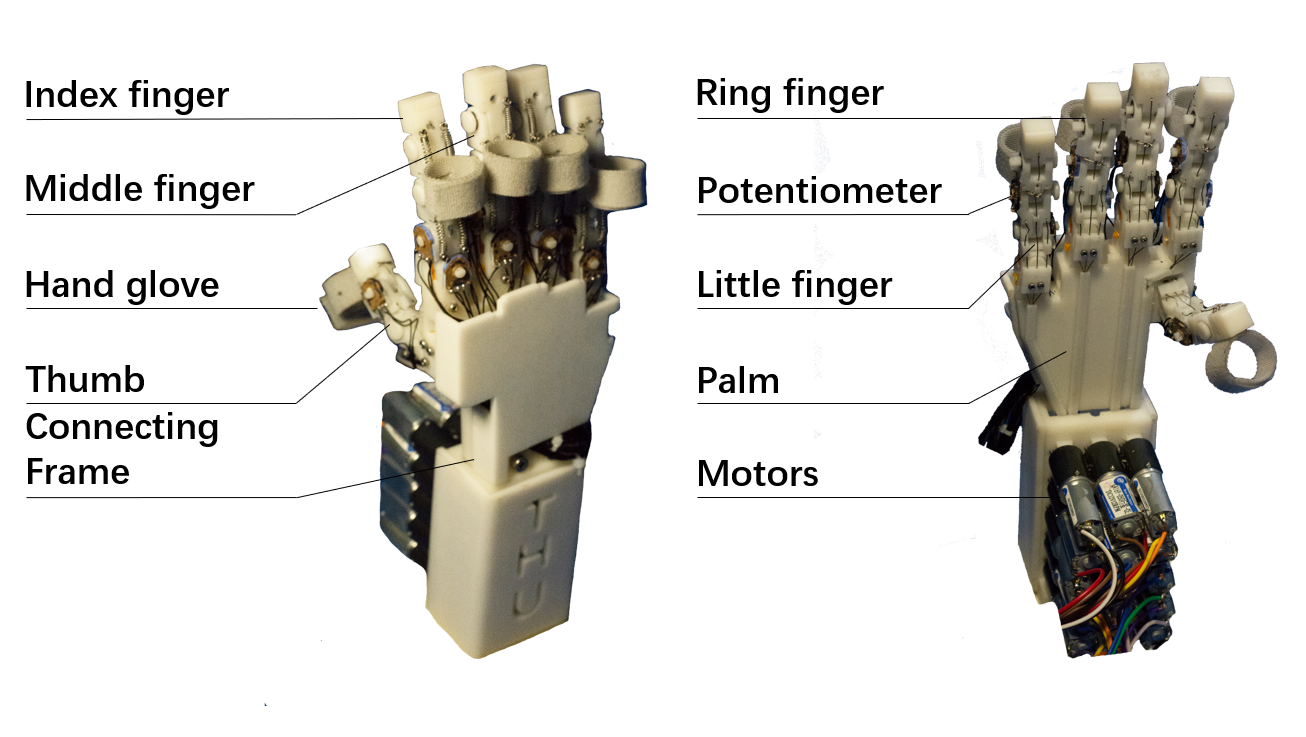}
    \caption{The overview of \ourhand that comprises a hand glove, five mechanical fingers, fifteen motors, and a palm.}
    \label{fig2}
\end{figure}
\begin{figure}[ht]
    \centering \includegraphics[width=8.5cm]{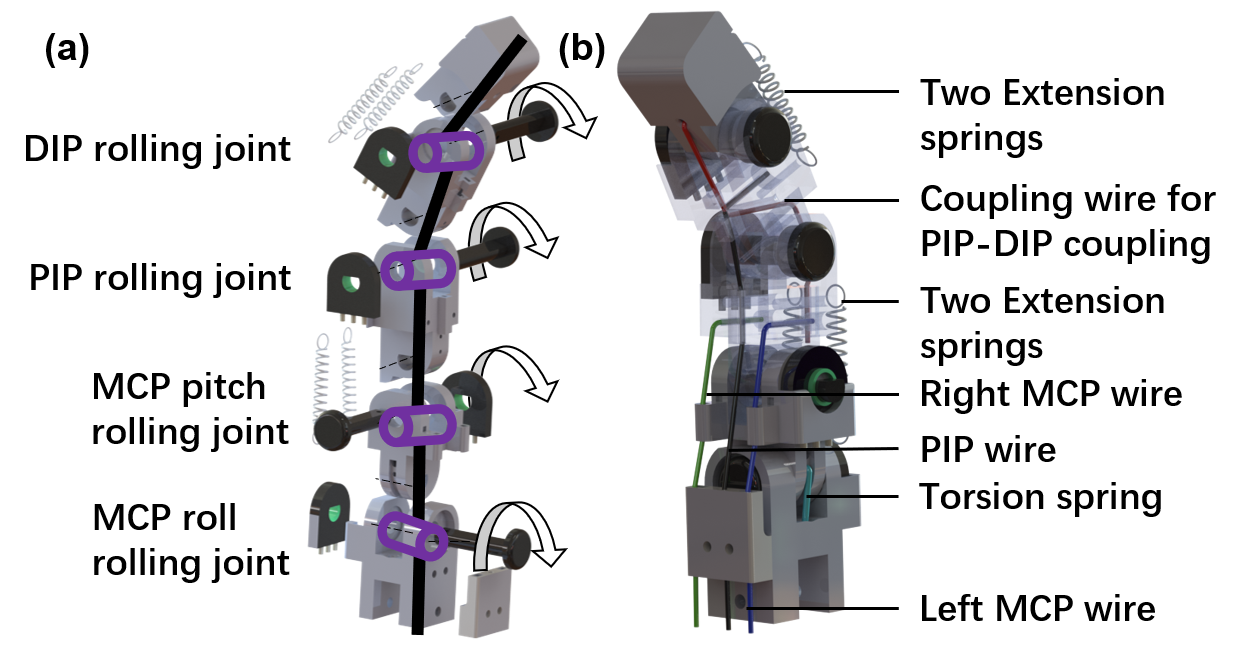}
    \caption{The middle finger structure of the HIRO Hand. (a) Exploded view of the single finger with the DIP rolling joint, PIP rolling joint, MCP pitch rolling joint and MCP roll rolling joint. (b) Perspective structure with driving wire details, including the coupling wire for DIP-PIP coupling, four extension springs, the torsion spring, the right MCP wire, the PIP wire and the left MCP wire. }
    \label{fig3}
\end{figure}
\subsection{Mechanical Design}
The mechanical system comprises a robotic hand, glove rings, and a driver module as shown in  Fig.~\ref{fig2}. 
The robotic hand is mainly 3D-printed and consists of a palm and five fingers. 
Each of the fingers has 3 DoFs. Thus, the \ourhand has 15 DoFs that can handle complex grasping or manipulation tasks. The overall design of the \ourhand adopts the structure of human hands~\citep{18}. Except for the thumb, each finger mimics the corresponding human finger and has four phalangeal bones: metacarpal, proximal, intermediate, and distal phalanges. Each finger also has three joints, namely the metacarpophalangeal~(MCP), the proximal interphalangeal~(PIP), and the distal interphalangeal~(DIP) joints, which have 2, 1, and 1 DoFs respectively.  The thumb shares the same design except for no intermediate phalanx. Therefore, we use the middle finger as an example in the following description. 
 As shown in Fig.~\ref{fig3} (a), the 3-DoF middle finger is 56.5 mm in length and has five 3D-printed rigid components. The length of each finger is usually shorter than that of an operator's hand, ensuring that the operator can wear the glove and operate the robotic hand smoothly. Each joint is integrated with a potentiometer that measures the real-time angle of each joint to the controller. 
In Fig.~\ref{fig3} (b), there are three groups of springs in each finger according to the distribution of finger extensors and Interosseous dorsal muscle. The first group consists of two springs that connect the DIP and PIP and the second group includes two springs connecting the PIP and MCP. Without external driving force, the springs in group one can retract the PIP and DIP joints, and the springs in group two can retract the MCP joint. Additionally, the MCP joint contains a torsion spring classified under Group 3 that possesses an inherent ability to restore the joint to its initial pose without the application of external force.
The tendons that enable joint rotation comprise four drive cables: the coupled wire, the PIP wire, the right MCP wire, and the left MCP wire. Fig.\ref{fig4} (a) illustrates the PIP-DIP coupling wire as a red curve, which connects the PIP and DIP joints. Additionally, the PIP wire is shown as a black curve in Fig.\ref{fig4} (a), which facilitates the flexion motion of the PIP joint. The yellow and orange wires in \ref{fig4} (b) are the left MCP and the right MCP wires that produce the MCP flexion and MCP adduction motion.
The glove rings are fasteners made of nylon so that human operators may wear the robotic hand. The fastener employs both hook and loop sides, allowing for a customized level of tightness through the manipulation of the intersection degree between the two. The fastener possesses a width of 10 mm and accommodates a range of human finger sizes with inner diameters spanning from 0 mm to 25.5 mm.
These attributes collectively contribute to the adaptability of the fastener, making it a suitable option for a variety of users.
The driver module is composed of 15 planetary reduction motors integrated with 15 3D-printed axle sleeves, with each sleeve connecting to a driving tendon that transmits the driving force to each joint. The driving tendon, which is composed of bundled wires, has a diameter of 0.4 mm. The motor weighs 55 g and is rated for a torque of $0.784$ N$\cdot$m. In the following section, we will detail about the bending angles of PIP, DIP, and their relationship.
\subsection{Joint Transmitting Model of PIP-DIP Coupling Structure}
\begin{figure}[t!]
    \centering
    \includegraphics[width=8.5cm]{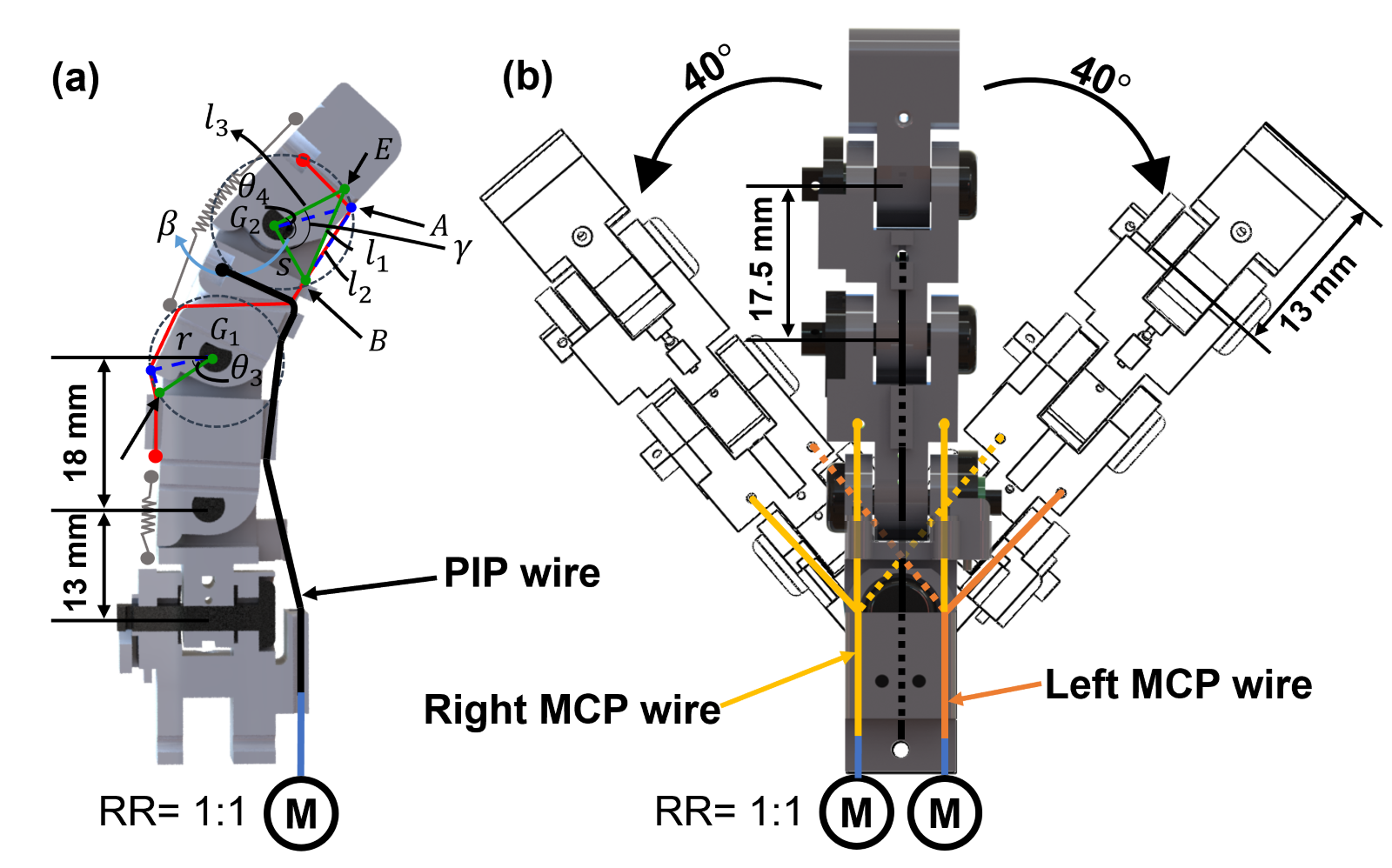}
    \caption{(a) Section view of the middle finger and definitions of parameters of the coupling wire (red) and PIP wire (black). The end of the drive wire 1 is directly connected to the motor with a reduction ratio of 1:1 and pulls the joint PIP directly. (b) Schematic diagram of the MCP rolling joint and the MCP pitch rolling joint actuation. The MCP right wire (yellow wire) and MCP left wire (orange wire) are shown. Simultaneously driving both cables to the same length causes the MCP pitch rolling joint to bend, while employing Differential Drive achieves flexion in the MCP rolling joint. The maximal bending angle achievable for the MCP rolling joint is 40 degrees. (RR: reduction ratio)}
\label{fig4}
\end{figure}

In this paragraph, we investigate the rotation angles of the PIP and DIP joints and their relationships. Specifically, we consider the scenario where the finger bends from an initial state, during which the PIP joint rotates by an angle of $ \theta_3 $ and the DIP joint rotates by an angle of $ \theta_4 $, as shown in Fig.\ref{fig4} (a). 
The wire on the right shown in Fig.~\ref{fig4} (a) represents the coupling wire. The turning point of the coupling wire at the DIP is point A in the same figure. We also mark point E as the initial state of point A when the finger is fully expanded. Point B on the PIP is where the coupling wire first crosses the PIP from the DIP above, and point C is located at the edge of the PIP. Points C and D will be the same points when the mechanical finger is fully expanded. G1 and G2 denote the center points of rotation of the PIP and DIP joints, respectively. 
To better illustrate the kinematic and geometric properties of \ourhand, we define all the notations first. The distance between point C and point D after the PIP joint rotates by an angle $ \theta_3 $ is denoted as $ l $, while the distance between point B and point E is denoted as $ l_1 $. 
The distance between point B and point A when the DIP joint is rotated by an angle of $ \theta_4 $ is denoted as $ l_2 $, and the distance between point E and point $G_2$ is denoted as $ l_3 $. The distance between point B and point $G_2$ is denoted as $ s $, while the angle between line segments A$G_2$ and B$G_2$ is denoted as $ \beta $ and the angle between E$G_2$ and B$G_2$ is denoted as $ \gamma $. The distance between point $G_1$ and point C is denoted as $ r $. 
Then, given the law of cosine, the length of $ l $ can be expressed as equations involving $ \theta_3 $ and $ r $. Furthermore, $ l_1 $ is equal to the sum of $ l $ and $ l_2 $, and $ \gamma $ is the sum of $ \theta_4 $ and $ \beta $. The relationship between $ \theta_3 $ and $ \theta_4 $ is calculated as:
\begin{equation}\label{eq1}
{\theta _4} = \gamma  - \arccos \left( {\frac{{{c_1} + 2{l_1}\sqrt {2{r^2}(1 - \cos {\theta _3})}  + 2{r^2}\cos {\theta _3})}}{{2{l_3}s}}} \right)
\end{equation}

where, 
\[\begin{array}{l}
{c_1} = l_3^2 + {s^2} - l_1^2 - 2{r^2}
\end{array}\]
\begin{figure}[t!]
    \centering
    \includegraphics[width=8.5cm]{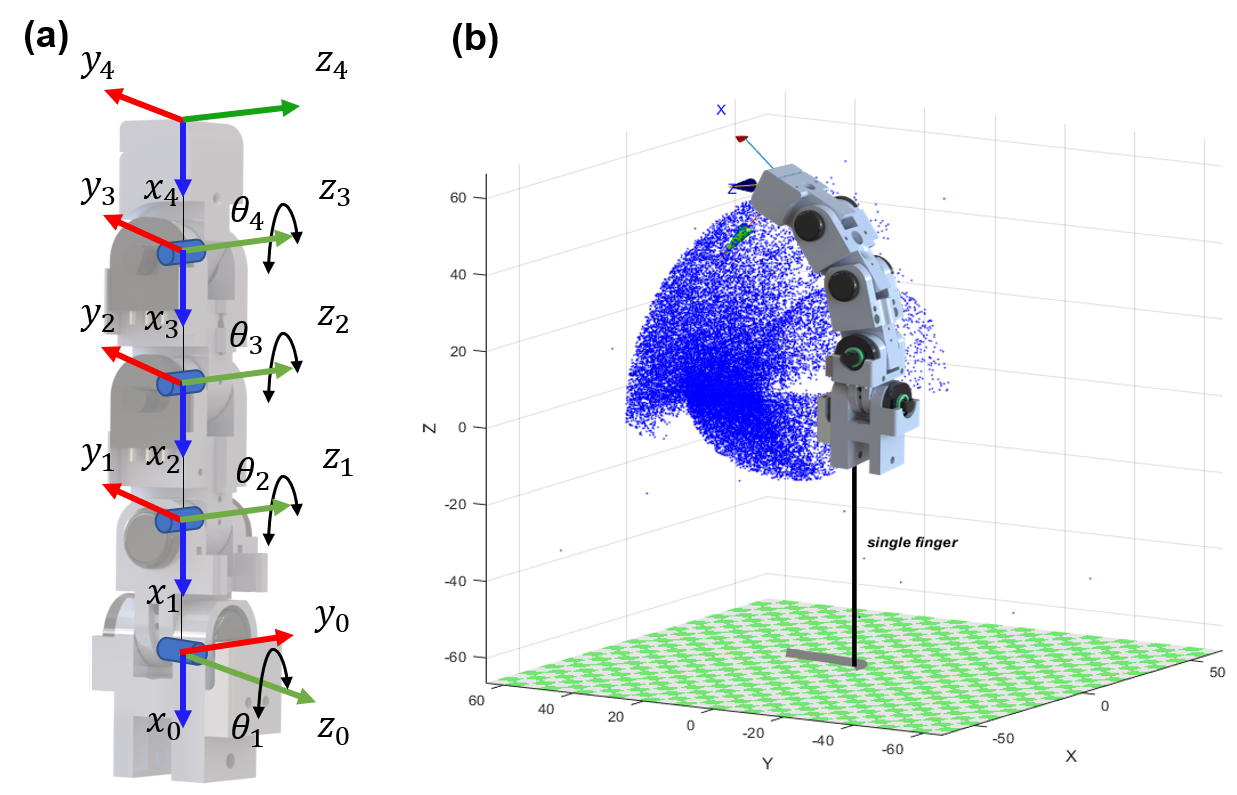}
    \caption{Coordinate system definition and workspace analysis of the middle finger of the HIRO Hand. (a) Coordinate system definition for forward kinematic. (b) Workspace of the middle finger obtained via the Monte Carlo tree search algorithm. The DIP, PIP, and MCP pitch joints are set to a roll of $ 90^{\circ} $, while the MCP rolling joint is set to a rotation of $ 180^{\circ} $, with $ 90^{\circ} $ to the left and $ 90^{\circ} $ to the right.}
\label{fig5}
\end{figure}
\begin{figure}[ht]
    \centering
    \includegraphics[width=8.5cm]{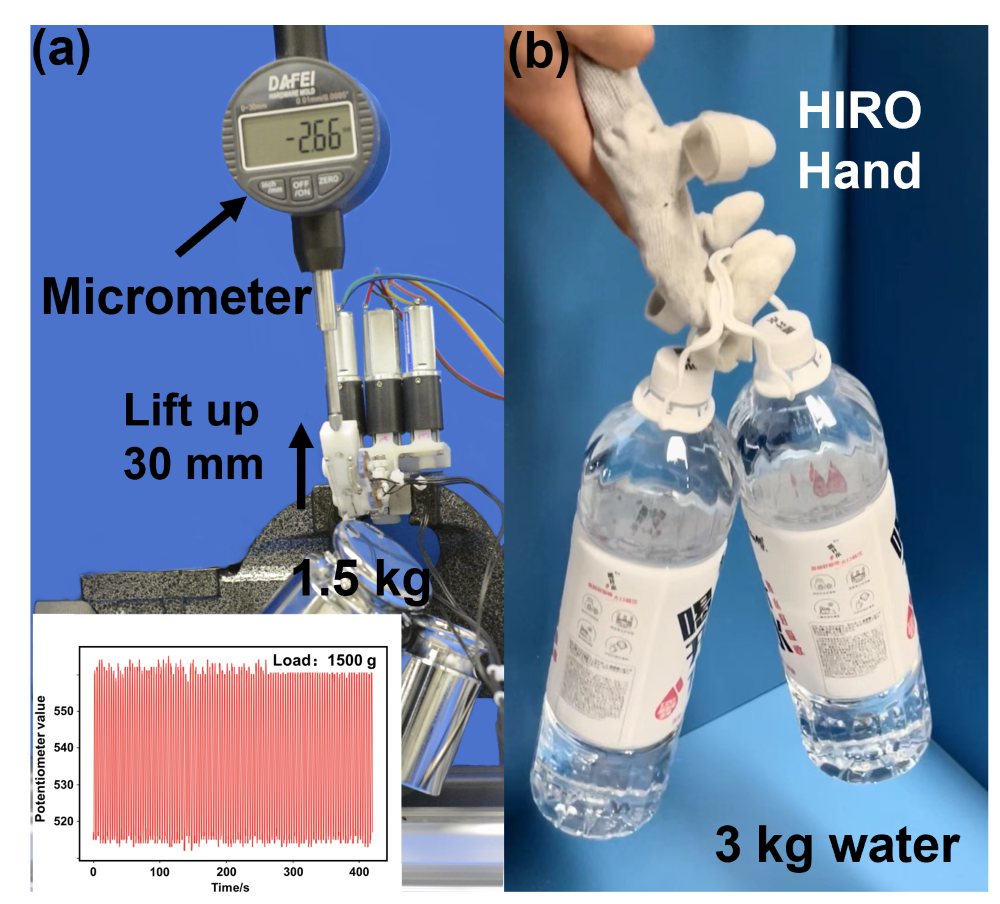}
    \caption{(a) Setup of the repeatability experiment, in which a load of $1.5$ kg is attached to the end tip of the finger, and the finger is commanded to lift the weight repeatedly over 100 times. (b) Performance of \ourhand in lifting 3 kg of water using three fingers.}
    \label{fig6}
\end{figure}
\begin{figure*}[htp]
    \centering
    \includegraphics[width=17cm]{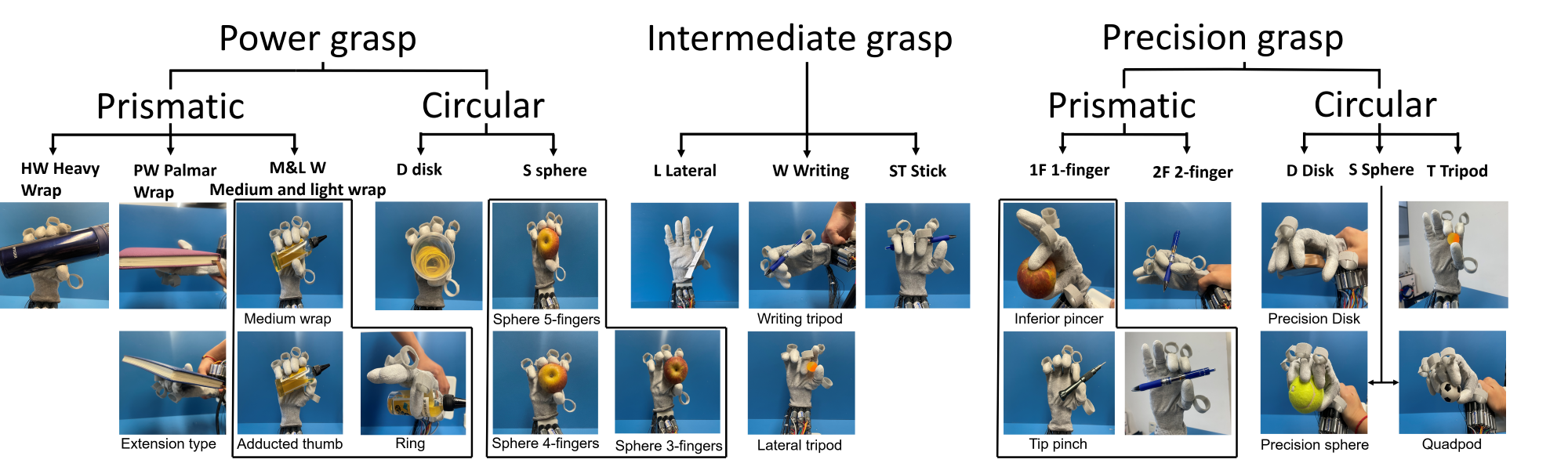}
    \caption{The hand taxonomy realized by the HIRO Hand. The \ourhand demonstrates the ability to perform 21 distinct grasp types, which collectively encompass nearly 0.80  of all grasp types in \citep{Cini_2019}.}
    \label{fig7}
\end{figure*}
 \begin{figure}[ht]
     \centering
     \includegraphics[width=8.5cm]{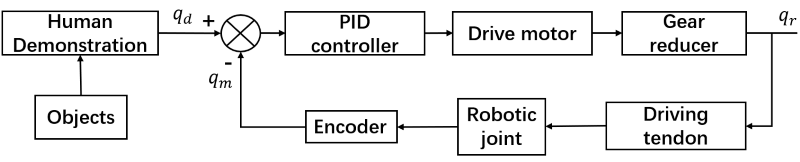}
     \caption{The overall scheme of the traditional controller for robot learning. The first step involves the manual demonstration of hand-over-hand movement by a human to the HIRO Hand. In the second step, the learned trajectory is transmitted to a PID controller to generate the desired joint positions, enabling the \ourhand to move to the target location.}
     \label{fig8}
 \end{figure}
 \begin{figure}[ht]
     \centering
       \includegraphics[width=8.5cm]{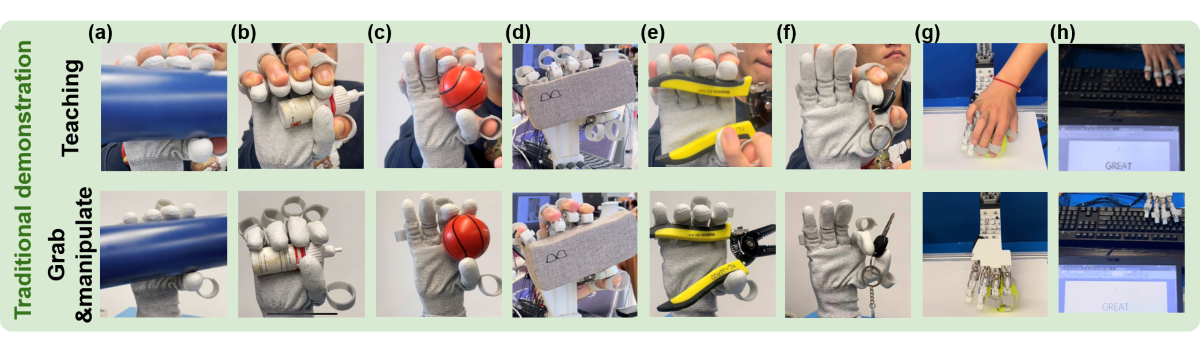}
     \caption{Overview of teaching the \ourhand to grasp different objects, including a cup, glue bottle, toy ball, tongs, and key, through manual demonstration. Additionally, in-hand manipulation skills, such as rotating a tennis ball and typing the word "GREAT" without any external assistance, are also demonstrated using hand-over-hand teaching based on traditional demonstration controllers.}
     \label{fig9}
 \end{figure}

  \begin{figure}[htp]
     \centering     
     \includegraphics[width=8.5cm]{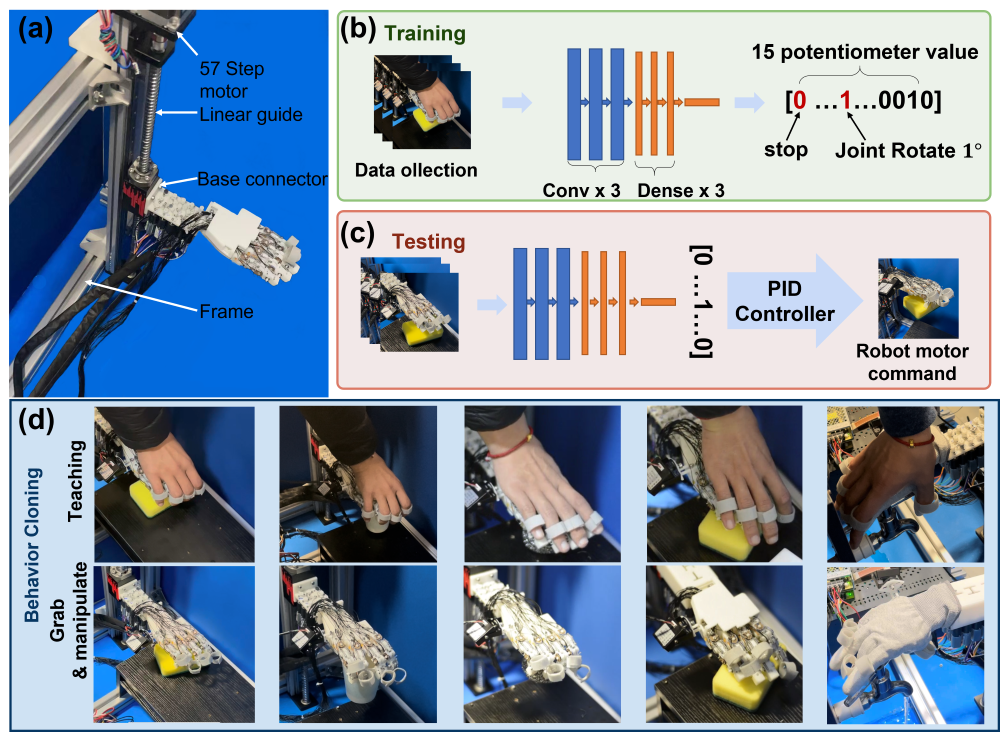}
    \caption{Experimental setup and training/testing process for the \ourhand. (a) The experimental setup comprises a step motor, a linear guide with a valid travel distance of $ 450 $ $ mm $, a base connector, the \ourhand, and a steel frame. (b) An overview of the training and testing process is presented, where hand-over-hand teaching videos and 15 potentiometer values are used as inputs and outputs of CNN. The output value is binary, with $ 0 $ representing a command to stop the motor and $ 1 $ representing a command to rotate the motor. (c) The grasping capabilities of the \ourhand are demonstrated for different handle types, including "egg crate" foam, a cup, and a wire ball, and in-hand manipulation tasks, such as rotating the egg foam and grasping and unscrewing a faucet, using the behavior cloning algorithm.
    }
    \label{fig10}

 \end{figure}

\subsection{Kinematics And Workspace}
\subsubsection{Forward Kinematic Analysis}
The primary objective of performing forward kinematic analysis is to compute the fingertip position of a robotic hand by leveraging the angles of its constituent joints. The kinematic model of the finger is determined through the use of the Denavit-Hartenberg (D-H) method~\cite{osti_5502022}. The D-H parameters are shown in Table \ref{tab:2}. The origin of the base frame, as illustrated in Fig.\ref{fig4} (a), is located at the MCP roll rolling joint.
Building upon the aforementioned discourse on the correlation between $ \theta_3 $ and $ \theta_4 $ in the previous section, we can streamline Eq.(\ref{eq1}) as follows:
\begin{equation}
    {\theta _4} = \lambda ({\theta _3})
\end{equation}

Thus, according to the Denavit-Hartenberg (D-H) method, the transmission matrix $T_0$, $T_1$, $T_2$ can be calculated. In order to illustrate the application of the D-H method in a standard manner, and to elucidate the representation of theta3 in terms of theta4, $T_3$ can be expressed as:

\begin{equation}
     {T_3} = \left[ {\begin{array}{*{20}{c}}
     {\cos (\lambda ({\theta _3}))}&{ - \sin (\lambda ({\theta _3}))}&0&{18\cos (\lambda ({\theta _3}))}\\
     {\sin (\lambda ({\theta _3}))}&{\cos (\lambda ({\theta _3}))}&0&{18\sin (\lambda ({\theta _3}))}\\
     0&0&1&0\\
     0&0&0&1
     \end{array}} \right] 
\end{equation}
The position and orientation of the fingertip with respect to the base coordinate system can be determined as follows:
\begin{equation}
     T_0^3 = {T_0} \cdot {T_1} \cdot {T_2} \cdot {T_3}
\end{equation}
\begin{table}[]
\centering
\caption{The Middle Finger D-H Parameter of The \protect\ourhand}
\begin{tabular}{cccccccccc}
\hline
\toprule
axis & $ \theta $ & d & a & $ \alpha $ & axis & $ \theta $ & d & a & $ \alpha $ \\ 
\midrule
1  & $ \theta_1 $ & 0 & 13 & $ \pi/2 $ & 2  & $ \theta_2 $ & 0 & 18 & 0 \\ 

3  & $ \theta_3 $ & 0 & 17.5 & 0 & 4  & $ \theta_4 $ & 0 & 18  &  0\\ 
\bottomrule
\end{tabular}
\label{tab:2}
\end{table}

\begin{table}[]
\centering
\caption{Repeatability Experiment Results with Varied Payloads}
\begin{tabular}{ccccc}

\toprule
Payload(g) & Times & Std.Deviation(mm) & Max Deviation(mm)  \\ 
\midrule
600  & 108 & 0.14 & 0.32  \\ 
900  & 117 & 0.08 & 0.16  \\ 
1200  & 110 & 0.13 & 0.36  \\ 
1500  & 100  & 0.11 & 0.26   \\ 
\bottomrule
\end{tabular}
\label{tab:3}
\end{table}
As depicted in Fig.~\ref{fig4} (b), the workspace is computed using a Monte Carlo tree search algorithm. In the current experiment, the pitch joints of DIP, PIP, and MCP are rolled up to $90^{\circ}$, while the MCP's rolling joint is rotated to $180^{\circ}$ with a $90^{\circ}$ inclination to both the left and right sides.
\subsubsection{End Effector Vector} 
The determination of the position and orientation of a robot's end effector is important as it interacts with the environment directly. 
We use end effector vector, which represents the end effector's location and orientation in a three-dimensional space, to control the end effector. The end effector vector can be expressed as:
\begin{equation}
s = \left[ {\begin{array}{*{20}{c}}
{{\sigma _3} + \mu_1\cos {\theta _3}{\sigma _1} - \mu_1\sin {\theta _3}{\sigma _2} + \mu_2}\\
0\\
{{\sigma _4} - \mu_1\cos {\theta _3}{\sigma _2} - \mu_1\sin {\theta _3}{\sigma _1}}\\
0\\
{{e^{{\theta _4}/\pi }}}\\
0
\end{array}} \right]
\end{equation}
where, $\mu_1$=18, $\mu_2$=13,
\[\begin{array}{l}
{\sigma _1} = \cos {\theta _1}\cos {\theta _2} - \sin {\theta _1}\sin {\theta _2} , \\ \\
{\sigma _2} = \cos {\theta _1}\sin {\theta _2} + \sin {\theta _1}\cos {\theta _2} , \\ \\ 
{\sigma _3} = 18\cos {\theta _1} + \frac{{35\cos {\theta _1}\cos {\theta _2}}}{2} - \frac{{35\sin {\theta _1}\sin {\theta _2}}}{2} , \\ \\
{\sigma _4} =  - 18\sin {\theta _1} + \frac{{35\cos {\theta _1}\sin {\theta _2}}}{2} - \frac{{35\sin {\theta _1}\cos {\theta _2}}}{2}
\end{array}\]
\section{Performance of The \ourhand \quad}
In this section, we test the performance of the \ourhand hand: 1) repeatability with different payloads; 2) the grasping ability; 3) the in-hand manipulation ability.
\subsection{Single Finger Repeatability with a Payload}
As depicted in Fig.~\ref{fig5} (a), a weight is attached to the fingertip, with a micrometer positioned beyond it. The weight is lifted up 30 mm and subsequently lowered to its original position within a single period. In the experiments, the finger is commanded to perform 100 periods while carrying varying payloads ranging from 500 g to 1500 g. The  driving force can then be assessed. Fig.~\ref{fig5} (b) illustrates that a finger of \ourhand is lifting a 3 kg bottled water using three fingers. Table \ref{tab:3} presents the standard and maximum deviations observed under different experimental conditions, with payloads ranging from 600 g to 1500 g. The results show a single finger is capable of lifting weights exceeding 1500 g with a standard deviation of 0.11 mm and a maximum deviation of 0.26 mm across 100 lifting cycles. Additionally, Fig.~\ref{fig6} illustrates the potentiometer value variation over time during the repeatability experiment.
\subsection{The Grasping Performance}
Grasping is a crucial operation for a robotic hand to perform later dexterous manipulation tasks. Grasping types are roughly categorized into three main groups: power grasps, intermediate grasps, and precision grasps. Within these categories, there are in total 28 fine-grained grasp types~\citep{Cini_2019}.
As demonstrated in Fig.~\ref{fig7}, the \ourhand's grasping ability is evaluated without any external assistance, and the findings reveal that the \ourhand is capable of executing 21 grasp types, covering nearly 0.80 of all grasp types.
\section{HAND OVER HAND TEACHING RESULTS}
In this section, we use both control-based and learning-based controllers to enable grasping and in-hand manipulation of objects from demonstrations using HIRO Hand.
\subsection{Teaching the to \ourhand Manipulate through PID-based Method}
We describe both the demonstration collection and the control algorithm in this section. As shown in  Fig.~\ref{fig8}, there are two steps. The first step is that, for different objects, a human operator wears the robotic hand as a glove and performs the desired task while each joint angle position $ q_d $ is being recorded. In the second step, a PID position controller~\cite{thomas2009position} is used, and given a desired position $ q_d $, the robotic hand reaches the desired position. The PID controller is expressed as follows:
 \begin{equation}
     e[i] = {q_m}[i] - {q_d}[i]
 \end{equation}
 \begin{equation}
     {q_r}[i] = {k_p}[i] \cdot e[i] + {k_i} \cdot \sum\limits_0^t {e[i]}  + {k_d} \cdot \frac{{(e[i] - pre\_e[i])}}{{dt}}
 \end{equation}
 where $ i $ represents the fifteen potentiometer values($k_p = 0.5$, $k_i = 0.1$). Here, $ q_d $ denotes the desired joint angle position, $ q_m $ denotes the feedback real-time joint angle position of each joint, and $ q_r $ represents the PWM output command for driving the motors. The term $ e[i] $ denotes the error between the real-time joint angle and the desired joint angle, while the value of $ pre_e[i] $ is the value of $ e[i] $ from the previous time step.  To evaluate the algorithm's performance, a series of experiments are conducted, which include grasping objects such as a cup, glue bottle, toy ball, class box, tongs, and a key, as shown in Fig.\ref{fig9} (a)-(e). Furthermore, the \ourhand is taught to rotate a tennis ball (Fig.\ref{fig9} (f)) and type the letter "GREAT" on a standard 433 $ \times $ 125 $ \times $ 30 mm keyboard (Fig.~\ref{fig9} (g)) without any external assistance, thereby demonstrating its advantage from multi-DOF flexibility.
 \subsection{Teaching the \ourhand to Manipulate with Visual Imitation Learning}
\begin{table}[]
\centering
\caption{Parameters of The Behavior Cloning algorithm(DL: Dense layers, CL: Convolution Layers)}
\begin{tabular}{ccccc}
\hline
\toprule
Output shape  & Activate function & Optimizer & CL  & DL\\ 
\midrule
15 & relu & Adam & 3 & 3 \\ 
\bottomrule
\end{tabular}

\label{paras}
\end{table}
\begin{table}[]
\centering
\caption{Comparison of common cable-driven robotic hands}
\begin{tabular}{ccccc}
\hline
\toprule
Name & Price & Power grasp & repeatability  & Dofs  \\
\midrule
Shadow hand lite& 60000 & 4 & N$/$A & 13  \\
Fllex hand lite& 60000 & N$/$A & 0.20 & 15  \\ 
Ours & 400 & more than 3 & 0.14 & 15  \\
\bottomrule
\end{tabular}

\label{price}
\end{table}
 We use the widely adopted behavior cloning as our imitation learning algorithm as shown in Fig.~\ref{fig10} (b)-(c). The \ourhand learns to perform grasping or manipulation tasks purely from visual inputs.
The \ourhand is fixed to a 1-DoF robotic arm via the base connector. The robotic arm comprises a 57mm stepper motor and a linear guide (FUYU Company) and is mounted on a steel frame, as illustrated in Fig.~\ref{fig10} (a). The robotic arm enables the \ourhand to execute unidirectional movement along the linear guide axis. The \ourhand is positioned $100$ $mm$ above the object and lowered until the object is in contact. Throughout this process. A USB industrial camera (HSK-200W, HSK Company) is fixed to the hand and captures fifteen images per second, while the corresponding fifteen potentiometer values are recorded as binary values (0 or 1). The fifteen values correspond to the fifteen joint angles, where 0 indicates no response is needed, and 1 indicates that the joint should rotate by $1^{\circ}$ using a PID controller. Following that, the training process employed convolutional neural networks(CNN)~\cite{CNN}. More parameters of the algorithm are shown in Table. \ref{paras}. Each training iteration consists of 45 steps, with a batch size of 75, and a total of 45 epochs is executed. The input images have a resolution of 160$\times$320 pixels. The RGB color space of the input images is converted to the YUV color space. Following this step, Gaussian blurring is applied to reduce image noise and details. Then, the image undergoes enhanced smoothing and normalization. Moreover, we perform luminance variation-based data augmentation techniques on the image, with a range from 0.2 to 1.2. Throughout each task, we process approximately 600 data samples gathered at a frequency of 30Hz. At test time, the \ourhand descends from the reference position, with concurrent real-time image capture by the camera. The CNN-based robot controller makes predictions for the output potentiometer values, which are subsequently transmitted to the PID controller($k_p = 0.5, k_i = 0.1$). The controller then generates the requisite motor commands for the robotic system. More details about the code are accessible through the link: \href{https://github.com/weier123/HIRO_HAND.git}{Code details}

Fig.~\ref{fig10} (c) illustrates the implementation of five experiments. The first three experiments tries to grasp the "egg crate" foam~(success rate: 0.60), cup~(success rate: 0.67), and wire ball~(success rate: 0.75), respectively. In these experiments, the objective of the \ourhand is to approach the target objects from a distance and perform grasping, which reflects scenarios where humans reach out to retrieve an object located some distance away. 
The fourth experiment involves rotating the "egg crate" foam~(success rate: 1.00). Notably, the \ourhand commences in close proximity to the foam to facilitate direct rotation. This experiment is designed to simulate real-world scenarios where individuals need to reorient objects as part of their daily activities.
The last experiment involves chained behaviors of grasping and manipulation of the faucet~(success rate: 0.75). More precisely, the proposed approach requires the \ourhand to perform 1) sub-task 1: approaching the faucet and performing grasping, followed by 2) sub-task 2: unscrewing the faucet before reaching an end. It is worth noting that the aforementioned sub-tasks are accomplished through an end-to-end behavior cloning algorithm, which effectively showcases the \ourhand's potential to execute sequential actions in real-world scenarios based on behavior cloning. The average success rate of the five experiments is 0.722 and the results reveal that the algorithm performs satisfactorily in the aforementioned tasks.

\section{CONCLUSIONS} 
In this paper, we present the \ourhand as a novel approach for data collecting and learning from human demonstrations, as well as presenting dexterous manipulation.
The developed finger can exert 15N fingertip force and has a 0.14mm average repeatability of fingertip position under multiple loads from 500g to 1500g. The \ourhand is completely fabricated using 3D printing technology, resulting in a cost below \$400. The comparison of common cable-driven robotic hands is shown in Table. \ref{price}.
Additionally, both PID- and imitation learning-based robot controllers are developed. Notably, the \ourhand is capable of accomplishing complex tasks such as sequentially grasping and unscrewing a faucet using an off-the-shelf behavior cloning algorithm. 
Future work involves collecting large-scale data by wearing this glove during daily activities, integrating more sensors to capture multi-modal signals from the environment, and integrating the \ourhand into a robot arm to undertake more intricate tasks. 

\bibliographystyle{unsrt}
\bibliography{refer}
\end{document}